\documentclass[11pt,a4paper]{article}

\usepackage{a4wide} 

\usepackage[utf8]{inputenc}
\usepackage[T1]{fontenc}
\usepackage{graphicx}

\newcommand\cad{c'est-à-dire } 

\newcommand\type[1]{^{#1}}
\newcommand\et\land
\newcommand\fl{\rightarrow} 

\newcommand\systF{\textsf{F} }
\newcommand{\ltyn}{\ensuremath{\Lambda\mathsf{Ty}_n}}

\newcommand\Land{\&^\Pi}

\newcommand\iepsilon{\epsilon^1}

\newcommand\flex{ \textsc{(f)}}  
\newcommand\rig{ \textsc{(r)}}

\newcommand\ttt{\mathbf{t}}
\newcommand\eee{\mathbf{e}}

\newcommand\event{\mathbf{v}}
\newcommand\ma[1]{"\emph{#1}"}

\newcommand\paragraphe[1]{\par\noindent\textbf{#1}\quad}

\usepackage[french]{babel}

\usepackage{fancyhdr} 
\usepackage{fancyhdr, graphicx}

\title{Sémantique des déterminants dans un cadre richement typé}

\author{Christian Retoré\\
\small  IRIT \& Université de Bordeaux\\ 
  \texttt{Christian.Retore@irit.fr} \\ 
}

\newcommand\citet\cite

\usepackage{gb4e}

\usepackage{hyperref}

\begin{document}

\maketitle
\thispagestyle{fancy} 

\paragraphe{Résumé:} 
La variation du sens des mots en contexte nous a conduit à enrichir le système de types utilisés dans notre analyse syntaxico-sémantique du français basé sur les grammaires catégorielles et la sémantique de Montague (ou la lambda-DRT). L’avantage majeur d’une telle sémantique profonde est de représenter le sens par des formules logiques aisément exploitables, par exemple par un moteur d’inférence. Déterminants et quantificateurs jouent un rôle fondamental dans la construction de ces formules, et il nous a fallu leur trouver des termes sémantiques adaptés à ce nouveau cadre. Nous proposons une solution inspirée des opérateurs epsilon et tau de Hilbert, éléments génériques qui s'apparentent à des fonctions de choix. 
Cette modélisation  unifie le traitement des différents types de déterminants et de quantificateurs et autorise le liage dynamique des pronoms. Surtout,  cette description  calculable des déterminants 
s’intègre parfaitement à l’analyseur à large échelle du français Grail, tant en théorie qu’en pratique.\\

\paragraphe{Mots-clés:}{Analyse sémantique automatique; Sémantique formelle; Compositionnalité;  }

\vfill 

\begin{center}
\Large \bf On the semantics of determiners

 in a rich type-theoretical framework 
\end{center} 

\vfill 

\paragraphe{Abstract:}{On the semantics of determiners in a rich type-theoretical framework}{
The variation of word meaning according to the context led us to enrich the type system of our syntactical and semantic analyser of French based on categorial grammars and Montague semantics (or lambda-DRT).  The main advantage of a deep semantic analyse is too represent meaning by logical formulae that can be easily used e.g. for inferences. Determiners and quantifiers play a fundamental role in the construction of those formulae and we needed to provide them  with semantic terms adapted to this new framework. We propose a solution inspired by the tau and epsilon operators of Hilbert, generic elements that resemble choice functions. This approach unifies the treatment of the different determiners and quantifiers and allows a dynamic binding of pronouns. Above all, this fully computational view of determiners fits in well within the wide coverage parser Grail, both from a theoretical and a practical viewpoint. 
\\
}

\paragraphe{Keywords:} Automated semantic analysis; Formal Semantics; Compositional Semantics.

\newpage 


\section{Présentation}

Dans le cadre du traitement automatique des langues, on entend plus souvent parler de sémantique distributionnelle, de vecteurs de mots et de fréquences que de sémantique formelle ou compositionnelle. Certes, les approches quantitatives sont plus aisées à mettre en oeuvre et fournissent des outils efficaces mais elles ne répondent pas aux mêmes questions. Les approches quantitatives  sont fort utiles  en recherche d'information et en classification car elles permettent de dire \emph{de quoi parle}  une phrase, une page web, un texte.
En revanche elles ne disent pas ce qu'affirme le texte analysé, \emph{qui fait quoi}. 
Une phrase peut très bien nier quelque chose, et ainsi causer une erreur à un système de recherche d'information (cf. exemple \ref{ouragan}).  Il faut aussi savoir reconnaître les pronoms, pour répondre à des questions comme \emph{Geach était-il l'élève de Wittgenstein?} à partir du web où on ne trouve que l'exemple (\ref{Geach}).\footnote{Sauf mention contraire, nos exemples proviennent d'Internet.} 

\begin{exe}
\ex \label{ouragan} Mais vérification faite, ce n’était \textsc{pas} un \textsc{ouragan} qui était passé par là. 
\ex \label{Geach} Bien qu’\textsc{il n}’ait \textsc{jamais} suivi l’enseignement académique de \textsc{ce dernier}, cependant \textsc{il en}  éprouva fortement l’influence. 
\end{exe} 

Ainsi, l'analyse sémantique complète et profonde d'une ou plusieurs phrases reste une tâche pertinente dans le traitement automatique des langues. Ce processus est utilement complété par des techniques statistiques, par exemple pour trouver les paragraphes pertinents ou pour définir des préférences contextuelles lorsque plusieurs sens sont possibles. Une fois 
les formules logiques construites à partir du texte, 
un moteur d'inférence est capable de dire si une formule ou un énoncé découle de ce qui a été analysé
--- plus facilement qu'avec des graphes sémantiques. 

Les déterminants (voir par exemple \cite{Corblin2004det}) sont un ingrédient important de l'analyse logique d'une phrase.  Les déterminants indéfinis correspondent généralement  à une quantification existentielle ou, ce qui s'en approche,  à l'introduction d'un référent de discours ---  il peuvent aussi exprimer une propriété notamment en position d'attribut, mais nous ne parlerons pas de ce dernier sens. 
Les déterminants définis correspondent  plutôt 
à la désignation d'un élément saillant du contexte, mais il arrive qu'ils introduisent un élément de discours. 
Assez souvent,  ils expriment la quantification vague comme \ma{beaucoup, la plupart}. 
La quantification universelle est assez rare, et elle peut s'exprimer par un usage générique de \ma{un, le, la,  les}.  

D'un  point de vue pratique, ce travail se situe dans le cadre des grammaires catégorielles 
qui sont une approche de la syntaxe très orientée vers la sémantique compositionnelle. En effet, il est assez aisé de déduire de la structure syntaxique proposée par  une grammaire catégorielle une représentation du sens sous forme logique. D'ailleurs, à notre connaissance,  les deux seuls systèmes produisant une analyse sémantique complète comme une formule logique (une DRS, en fait) sont basés sur les grammaires catégorielles: 
\emph{Boxer} de \citet{Bos2008STEP2} 
analyse de l'anglais par des \emph{Categorial Combinatory Grammars} 
tandis que nous utilisons \emph{Grail} de  \citet{moot10semi,moot10grail}
basé sur les  \emph{Multimodal Categorial Grammars}.  
Dans un cas comme dans l'autre,  la grammaire est acquise automatiquement sur corpus annoté. 
L'acquisition automatique de la grammaire produit un grand nombre de catégories par mot, et un minimum de traitement probabiliste est nécessaire pour ne considérer que  les assignations les plus probables lors de l'analyse. Du point de vue sémantique, ces systèmes utilisent la correspondance entre syntaxe et sémantique telle qu'initiée par Montague c.f. \cite[chapitre 3]{MootRetore2012lcg}. Rappelons la brièvement sur un exemple, car c'est le point de départ de nos travaux. 

Supposons que l'analyse syntaxique de  "\emph{un club a\_battu Leeds.}" produise 
\ma{(un\ (club)) (a\_battu\ Leeds)} 
expression dans laquelle  la fonction est systématiquement écrite  à gauche. 
Si les termes sémantiques sont ceux du lexique de la figure 
 \ref{semanticlexicon}, alors 
 en remplaçant les mots par les termes sémantiques associés on obtient un grand $\lambda$-terme, que l'on peut réduire: 
\begin{figure} 
\begin{center} 
\begin{tabular}{ll} \hline 
\textbf{Mot} &  \textbf{\itshape type sémantique $u^*$}\\ 
& \textbf{\itshape  terme semantique~: $\lambda$-terme de type $u^*$}\\ 
&  {\itshape  $x\type{v}$ la variable ou la constante $x$ 
est de type  $v$}\\ \hline 
\textit{un} 
& $(e\fl t)\fl ((e\fl t) \fl t)$\\ 
& $\lambda P\type{e\fl t}\  \lambda Q\type{e\fl t}\  
(\exists\type{(e\fl t)\fl t}\  (\lambda x\type{e}  (\et\type{t\fl (t\fl t)} (P\ x) (Q\ x))))$ \\  \hline 
\textit{club}  & $e\fl t$\\ 
& $\lambda x\type{e} (\texttt{club}\type{e\fl t}\  x)$\\  \hline 
\textit{a\_battu} & $e\fl (e \fl t)$\\ 
& $\lambda y\type{e}\  \lambda x\type{e}\  ((\texttt{a\_battu}\type{e \fl (e \fl t)}\  x)  y)$ \\  \hline 
\textit{Leeds} &$e$ \\ &  Leeds 
\end{tabular}
\end{center} 
\caption{Un lexique sémantique élémentaire} 
\label{semanticlexicon}
\end{figure} 

$$
\begin{array}{c} 
\Big(\big(\lambda P\type{e\fl t}\ \lambda Q\type{e\fl t}\  (\exists\type{(e\fl t)\fl t}\  (\lambda x\type{e}  (\et (P\ x) (Q\ x))))\big)
\big(\lambda x\type{e} (\texttt{club}\type{e\fl t}\  x)\big)\Big) \\ 
\Big(
\big(\lambda y\type{e}\  \lambda x\type{e}\  ((\texttt{a\_battu}\type{e\fl (e\fl t)}\  x)  y)\big)\ Leeds\type{e}\Big)\\ 
\multicolumn{1}{c}{\downarrow \beta}\\ 
\big(\lambda Q\type{e\fl t}\  (\exists\type{(e\fl t)\fl t}\  (\lambda x\type{e}  (\et\type{t\fl (t\fl t)}  
(\texttt{club}\type{e\fl t}\  x) (Q\ x))))\big)\\ 
\big(\lambda x\type{e} \ ((\texttt{a\_battu}\type{e\fl (e \fl t)}\  x)  Leeds\type{e})\big)\\ 
\multicolumn{1}{c}{\downarrow \beta}\\ 
\big(\exists\type{(e\fl t)\fl t}\  (\lambda x\type{e}  (\et (\texttt{club}\type{e\fl t}\  x) ((\texttt{a\_battu}\type{e\fl (e\fl t)}\  x)  Leeds\type{e})))\big)
\end{array}
$$

Ce  $\lambda$-terme de type $\ttt$ 
peut être appelé la \emph{forme logique de la phrase}; il  
est plus agréable sous un format standard: 
$\exists x:e\  (\texttt{club}(x)\ \et\ \texttt{a\_battu}(x,Leeds))$.
Nous verrons ci-après que ce traitement standard de la quantification pose problème, notamment avec la sémantique lexicale. 

On observera qu'il y a deux logiques à l'oeuvre. La première est le calcul propositionnel intuitionniste dont on n'utilise que les preuves ou $\lambda$-termes: elle assemble des formules partielles. La seconde est une logique dont on n'utilise que les formules. Le $\lambda$-terme de type $\ttt$ obtenu \emph{in fine} est effectivement une formulation logique du sens en $\lambda$-calcul. 

\section{Déterminants et quantificateurs} 

Sans surprise, les déterminants considérés  sont de deux sortes,  définis et indéfinis --- nous essaierons d'éviter les pluriels: ils posent d'autres problèmes 
abordés dans ce cadre par \citet{MootRetore2011coconat}. 
Logiquement, les déterminants indéfinis s'apparentent à une quantification existentielle dite généralisée lorsqu'il agissent sur une classe ou lorsqu'ils introduisent un nouveau référent de discours --- nous n'aborderons pas le cas où ils introduisent une propriété, par exemple dans un groupe nominal attribut. 
Précisons de suite qu'un travail de formalisation et d'automatisation comme le nôtre  ne peut 
prétendre atteindre la finesse de travaux plus descriptifs comme ceux de 
\citet{Corblin2004det}, et que nous sommes donc contraints de schématiser, voire d'ignorer, certaines constructions. 
Considérons quelques exemples:
\begin{exe}
\ex  \label{unpiedunparent} 
\begin{xlist}
\ex \label{unpied} 
J'ai senti \textbf{un} animal \textsc{me toucher le pied}.
\ex \label{unparent} 
\textbf{Un} parent d'élève de maternelle \textsc{vient chercher son enfant} en état d'ébriété,\footnote{Il s'agit sans doute du parent :-)} l'enseignant commet-il une faute en remettant l'enfant à ce parent? 
\end{xlist} 
\ex \label{qqchpiedqqunenfant}
\begin{xlist}
\ex \label{qqchpied} 
Aujourd'hui, je me suis réveillé en sursaut parce que j'ai senti \textbf{quelque chose} \textsc{me toucher le pied}. Il s'est avéré que c'était mon autre pied.
\ex \label{qqunenfant} 
Précisez si \textbf{quelqu’un} \textsc{vient chercher l’enfant}. 
\end{xlist} 
\ex \label{lanimalleparent} 
\begin{xlist} 
\ex  \label{lanimal} 
Il y avait \textbf{une panthère sortie de la cage}. \textbf{Elle} était attachée.  \textbf{L'animal} a sauté sur moi.
\ex \label{leparent} 
\textbf{Un homme} avait menacé la principale du collège de Monts où son fils était scolarisé. \textbf{Le parent d’élève} a été condamné hier.
\end{xlist}
\ex \label{spa} 
A la SPA si ont désire adopter \textbf{un animal} il faut donner 500F, et on a 24 Heures pour réfléchir si l'on désire \textbf{l'animal} ou non.
\end{exe} 

Les deux premiers exemples (\ref{unpiedunparent}) sont à mettre en parallèle avec les deux suivants 
(\ref{qqchpiedqqunenfant}) 
qui  correspondent eux-aussi à une quantification existentielle. Dans cette deuxième version,  il n'y a que le prédicat principal,  que nous avons choisi pour être le même, et il n'y a plus de restriction à une classe d'objets par un nom commun,  avec ou sans compléments ($\overline{N}$ syntaxiquement ou $\eee\fl\ttt$ sémantiquement). Observons que le traitement usuel de la quantification dans une logique non typée à la Frege ne fait aucun distinction entre le nom quantifié  et le prédicat principal.  

Les déterminants définis ont un rapport avec les déterminants indéfinis, qui souvent les introduisent, comme le montre les exemples 
(\ref{lanimalleparent})
Les expressions se correspondent, et idéalement on aimerait que \ma{le X} soit précédé de \ma{un X},
comme dans l'exemple (\ref{spa}).  
En fait, c'est plutôt rare,  et les exemples en corpus sont plutôt comme (\ref{lanimalleparent}): l'antécédent de l'anaphore associative n'est pas celui qu'on espèrerait pour un traitement automatique, quelques inférences sont nécessaires. 

\subsection{Traitement usuel et critique} 

L'analyse traditionnelle attribue à l'article indéfini un terme sémantique exprimant une quantification existentielle. 

\begin{exe}
\ex \label{uncg} 
un:
$\lambda P^{\eee\fl\ttt}\lambda Q^{\eee\fl\ttt}(\exists \lambda x^{\ttt}. \&(P\ x)(Q\ x)): (\eee\fl\ttt)\fl(\eee\fl\ttt)\fl\ttt$
\ex \label{qqchcg} quelque\ chose: $\exists:(\eee\fl\ttt)\fl\ttt$ 
\end{exe} 

Les articles définis sont traités différemment: les groupes nominaux qu'ils introduisent sont plutôt vus comme des anaphores, dites  associatives, dont on cherche les référents.\footnote{Une autre approche utilise une fonction de choix, mais nous allons justement présenter une solution de cet ordre.} Cette modélisation classique en sémantique formelle ou dans les grammaires catégorielles pose divers problèmes. 

\paragraphe{Syntaxe et sémantique} Les déterminants  traités comme des quantificateurs généralisés mettent à mal la correspondance entre syntaxe et sémantique. La structure sémantique (\ref{strsem})  et la structure syntaxique (\ref{strsynt}) ne coïncident pas. 
Les grammaires catégorielles obtiennent une structure syntaxique (\ref{strsem}), laquelle n'a rien de naturel, 
au prix  d'une catégorie syntaxique différente pour chaque position syntaxique du groupe nominal quantifié:  cela n'est guère satisfaisant.   
\begin{exe}
\ex 
\begin{xlist}
\ex 
elle écoutait une chanson de lassana hawa
\ex \label{strsynt}
\textsc{synt. usuelle:} (elle (écoutait (\textbf{une} (chanson (de lassana hawa)))))
\ex \label{strsem} 
\textsc{sem. \& cg:} ((\textbf{une} \underline{(chanson (de lassana hawa))}) \underline{($\lambda x$ elle écoutait $x$)}) 
\end{xlist} 
\end{exe}

\paragraphe{Référence du groupe nominal quantifié} Comme le fait remarquer \cite{geach1962reference},  on peut se forger une interprétation du groupe nominal défini ou indéfini, avant même que le prédicat principal soit énoncé... et dans l'exemple (\ref{luth})  il n'arrive jamais. 
\begin{exe}
\ex 
\begin{xlist} 
\ex Ensuite, \textsc{les élèves} sont allés en salle info, pour réaliser un caryotype classé.
\ex Ensuite, \textsc{des élèves} sont venus voir ce que l'on faisait.
\ex \label{luth} Un luth, une mandore, une viole, que Michel-Ange [...].  [phrase nominale].\footnote{Mathias Enard, \emph{Parle-leur de batailles, de rois et d'éléphants} Actes-Sud, 2010.} 
\end{xlist} 
\end{exe} 

\paragraphe{Asymétrie entre thème et rhème} 
L'approche standard impose une symétrie entre le prédicat principal et la restriction à une classe d'objets, symétrie que la langue ne fait pas. 
\begin{exe}
\ex 
\begin{xlist}
\ex Certains politiciens sont des menteurs car ce qui les intéresse (...) 
\ex 
* Certains menteurs sont des politiciens car ce qui les intéresse (...) 
\footnote{Cet exemple est de nous, pour faire contraste avec le précédent.} 
\end{xlist} 
\end{exe} 

\paragraphe{Définis et indéfinis} Comme remarqué dans \cite{EgliHeusinger1995,Heusinger1997,Heusinger2004},
l'unicité est loin d'être requise lorsque l'on utilise un déterminant défini. 
Un locuteur peut dire \ma{l'île} du lac de Constance, alors qu'il y en a trois, comme ici sur Internet: 
\begin{exe} 
\ex 
Recueilli (...) 
par les moines de l'abbaye de Reichenau, 
sur \textbf{l'ile du lac de Constance}, 
\end{exe} 
De plus, "un" et "le" se rapprochent aussi car le  contexte extra linguistique permet parfaitement d'utiliser l'article défini sans que le référent n'ait jamais été introduit. 
\begin{exe} 
\ex
J'avais pris l'assurance 'automatiquement' avec le prêt immobilier lors de l'achat de \emph{la maison}. \footnote{Dans ce récit trouvé sur une FAQ il n'a jamais été question de "maison" auparavant} 
\end{exe}
Les déterminants \ma{un}  et \ma{le} se ressemblent, alors que la sémantique formelle usuelle les opposent. Selon von Heusinger il s'agit d'une différence d'interprétation et non de forme logique:  \ma{un} choisit un nouvel élément, tandis que \ma{le} choisit le plus saillant des référents possibles. 

\paragraphe{Les pronoms de type E}  ne sont pas des pronoms particuliers, mais une interprétation possible et très naturelle des  pronoms due à  \citet{Evans77pronouns}. Cette interprétation consiste à associer au pronom le terme sémantique de son antécédent.
On peut aussi traiter de la sorte les groupes nominaux introduits par l'article  défini \cite{EgliHeusinger1995,Heusinger1997,Heusinger2004}. 
Cela permet d'étendre la portée du quantificateur existentiel  souvent introduit par \ma{un} 
comme le font la DRT et la \emph{dynamic predicate logic}. 
Ce type d'interprétation n'est pas possible avec le 
terme sémantique standard associé à \ma{un} en (\ref{uncg}). 
\begin{exe}
\ex \label{hommebis} 
\begin{xlist}
\ex 
Soudain, \textbf{un homme} est entré.
\ex  \textsc{Il / Cet homme / L'homme} a hurlé << Donne-moi la caisse ! >>.  
\end{xlist} 
\end{exe}

\section{Opérateurs de Hilbert, quantificateurs et déterminants} 
\label{Hilbert} 

Les opérateurs de Hilbert, surtout $\iota$ et $\epsilon$, ont été utilisés pour modéliser les déterminants et la quantification existentielle, 
 en particulier par  von Heusinger  \cite{EgliHeusinger1995,Heusinger1997,Heusinger2004}. 
Ces opérateurs, bien décrits par \citet{HBvol2}, 
s'apparentent aux fonctions de choix (2nd ordre) 
qui elles  mêmes  se rapprochent des fonctions de Skolem (1er ordre, la quantification sur ces fonctions étant reportée au moment de leur interprétation) --- voir par exemple 
\citet{Steedman2012scope}.  Mais les opérateurs de Hilbert ne sont pas comparables avec ces autres formes de quantification: ils incluent les quantificateurs usuels, mais permettent en outre une forme de liage dynamique (comme dans \emph{dynamic predicate logic}) ainsi que des dépendances  complexes à la manière des quantificateurs branchants de Henkin. Pour davantage de précision sur les opérateurs de Hilbert, on pourra consulter \citet{epsilonIEP} ou \citet{espilonURL}.

\citet{Russell1905} eut le premier l'idée d'introduire un \emph{terme} --- un individu --- noté $\iota_x P(x)$ comme interprétation logique d'une description définie \ma{le\ P} où \ma{P} est une propriété, une formule à une variable libre. Que dénote ce terme? Rien s'il n'existe pas un unique individu tel que $P(x)$, et sinon cet unique individu. Mais chacun sait que le quantificateur \ma{il existe un unique $x$ tel que $P(x)$}  
($\exists !x P(x)$) n'a pas de bonnes propriétés, notamment parce que sa négation, \emph{aucun ou au moins deux} n'a rien de naturel: $\lnot\exists!x.\ P(x)\equiv (\forall x\lnot P(x))\lor (\exists y \exists z (y\neq z)\& P(y)\land P(z))$. 

Hilbert a donc reformulé ces termes génériques en laissant de côté  la condition d'unicité. 
Il associe un terme $\epsilon_x F(x)$  à toute formule $F(x)$, qui permet d'exprimer la quantification existentielle puisque 
$F(\epsilon_x F(x))\equiv \exists x\ F(x)$. Il introduit aussi son dual  $\tau_x F(x)$ qui 
permet d'exprimer la quantification universelle par $F(\tau_x F(x))\equiv \forall x\ F(x)$.
Les opérateurs  $\epsilon_x$ et $\tau_x$ lient la variable $x$ dans $F(x)$. 
Bien sûr, au vu de cette dualité, un seul des deux opérateurs $\tau$ et $\epsilon$ suffit,  si on dispose de la négation. 
Il est très compliqué d'interpréter ces termes en toute généralité puisqu'on sort de la logique du premier ordre.  Les modèles correspondant sont très complexes, voire mal définis \cite{Asser1957}. En revanche, pour les formules du calcul de Hilbert qui correspondent à des formules habituelles, les modèles usuels fonctionnent et le théorème de  complétude est vérifié. 
En l'absence de modèles simples, définissons la ``vérité" de ces formules en termes de déduction, d'autant que 
les  règles de déduction définissant ces opérateurs sont 
les règles usuelles de la quantification: 
\begin{itemize}
\item 
De $F(t)$ où $t$ est n'importe quel terme, on peut déduire $F(\epsilon_x F(x))$ \cad  $\exists x\ F(x)$. 
\item 
Si l'on a établi  $F(x)$ sans rien supposer sur $x$, on peut en déduire $F(\tau_x F(x))$ \cad $\forall x\ F(x)$ 
\end{itemize}

Pour les applications linguistiques, seul  $\epsilon$ a été utilisé:  la quantification existentielle joue un rôle central dans la langue, par exemple la DRT organise le discours autour des quantifications existentielles.
L'idée véhiculée par cet $\epsilon$ est simplement de construire un terme générique associé au groupe nominal quantifié. Par exemple, pour \ma{un enfant sage} on forme le terme $\epsilon_x. (enfant(x) \& sage(x))$.
Selon von Heusinger, pour \ma{l'enfant sage} le terme est quasi identique: 
$\iepsilon_x. (enfant(x) \& sage(x))$.\footnote{Les notations de von Heusinger sont source de confusion. 
Il note $\eta$ notre $\epsilon$ qui correspond au  \ma{un} existentiel et qui s'interprète toujours par un nouvel individu, tandis qu'il note $\epsilon$ notre $\epsilon^1$ qui correspond à l'article défini \ma{le,la} sans contrainte d'unicité et qui est interprété par l'élément le plus saillant.}  La différence entre  $\iepsilon$ et $\epsilon$ n'est qu'une différence d'interprétation: $\iepsilon$ choisit le plus saillant  en contexte tandis que $\epsilon$ en choisit un nouveau. 
Le typage à la Montague de ces opérateurs n'est pas donné. Cependant, $\epsilon$ et $\iepsilon$ sont de type $(\eee\fl\ttt)\fl\eee$: $\epsilon$ et $\iepsilon$ produisent un individu (un terme) à partir d'une propriété (ici: $enfant(x) \& sage(x)$), comme le ferait une fonction de choix.

\section{Rappels sur le lexique génératif montagovien}

Dans  \cite{BMRjolli},  nous avons proposé un lexique syntaxique et sémantique qui étend considérablement la sémantique de Montague pour rendre compte de l'adaptation 
du sens d'un mot au contexte.  Ce modèle s'est avéré pertinent pour des questions de sémantique lexicale ou compositionnelle: coprédications possibles ou non, ambiguité des déverbaux \cite{RealCoelhoRetore2013unilog}, le voyageur fictif  \cite{MPR2011taln}, pluriels \cite{MootRetore2011coconat}, termes génériques \cite{Retore2012rlv}). Notre modèle est assez proche de \cite{asher-webofwords,Luo2011lacl,Luo2012lacl}, mais nous sommes les premiers à aborder la quantification et des déterminants dans 
un cadre adapté à la sémantique lexicale. 

La question initiale qui nous a conduit à utiliser une théorie des types plus sophistiquée que celle de Montague est fort simple: comment rendre compte  des restrictions de sélection? Plus concrètement,  comment rejeter les deux premiers exemples et accepter les suivants? 
\begin{exe} 
\ex * leur \textbf{dix} est \textsc{bon} \label{dix} [cf. ex. (\ref{rugby})] 
\ex 
\begin{xlist} 
\ex \label{chaise} * Une \textbf{chaise} \textsc{aboie} souvent. [ex. inventé]
\ex Mon \textbf{chiot} \textsc{aboie} souvent pour m'inciter à jouer avec. 
\end{xlist} 
\ex 
\begin{xlist} \ex \label{club} \textbf{Barcelone} \textsc{a battu} Benfica 2-0. [club]
\ex \label{institution} \textbf{Barcelone} (/*et) \textsc{a choisi de structurer le réseau routier} de manière à préserver un centre ville piétonnier. [institution] 
\end{xlist} 
\ex 
\begin{xlist} 
\ex \label{livreI} Mon premier \textbf{livre} de cuisine … Mon livre \textsc{fétiche} à cette époque !
\ex  \label{livreO} Je l’ai \textsc{retrouvé}, il y a peu, chez ma maman [mon premier livre de cuisine]
\end{xlist} 
\end{exe} 

En utilisant différents types d'entités et en spécifiant le type d'objet attendu par les prédicats, les compositions sémantiquement impossibles produisent  des conflits de types. Le sujet du verbe \ma{aboyer} doit être un chien, ou tout au moins un animal, (\ref{chaise}) un nombre ne saurait être \ma{leur} ni être \ma{bon} (\ref{dix}) etc. L'impossibilité sémantique est matérialisée par l'application d'un prédicat $P^{\xi\fl\ttt}$ présupposant un argument de type $\xi$ (par exemple \ma{animal}) à un argument $a^{\alpha}$ d'un autre type $\alpha$ (par exemple, \ma{meuble}) avec $\alpha\neq\xi$: 
$P^{\xi\fl\ttt} a^{\alpha}$

On notera qu'il faut parfois relaxer ces contraintes. Dans une discussion sur Internet au sujet 
d'un prochain  match de rugby,
l'exemple (\ref{dix}) ci-dessus se trouve:  
\begin{exe} 
\ex \label{rugby} 
si leur \textbf{dix} est \textsc{bon} ils nous torchent ça c'est sûr 
\end{exe} 

Il faut aussi prévoir que certaines coprédications sont heureuses --- (\ref{livreI} et \ref{livreO}) ---
et d'autres moins ---  (\ref{club}) et  (\ref{institution}) avec  \ma{et} à la place de Barcelone.  

Pour traiter tous ces phénomènes, nous avons proposé un lexique sémantique catégoriel où chaque mot se voit associer un $\lambda$-terme principal, qui ressemble beaucoup à celui de la sémantique de Montague rappelée ci-dessus, ainsi que des $\lambda$-termes optionnels qui permettent de transformer un mot dans l'aspect souhaité, par exemple un numéro en joueur de rugby. 
En raison du grand nombre de types, il convient de factoriser les opérations sur des termes de types différents et  d'avoir des opérations sur des familles de types, et nous nous sommes donc placés dans le  $\lambda$-calcul du second ordre appelé système \systF --- mais d'autres théories des types comme celle de \cite{Luo2012lacl} seraient également possibles.  En revanche, notre système se distingue surtout par le caractère lexical et non ontologique des transformations lexicales: celles-ci sont déclenchées par les mots et non par le type des mots. Cela nous semble pleinement justifié par des exemples comme \ma{promotion} et \ma{classe}: les deux désignent des groupes d'élèves, mais seul
 \ma{classe},  peut désigner un lieu (la salle de classe), tandis que \ma{promotion} ne le peut pas. 
Les coprédications possibles ou impossibles, dans ce système où les mots portent les transformations, sont modélisées en distinguant deux sortes de transformations : les transformations \emph{rigides}, qui imposent de ne référer qu'à cet aspect de l'objet, et les transformations \emph{flexibles}  qui permettent de renvoyer à un aspect de l'objet sans exclure les autres aspects dudit objet.

Notre cadre formel, le système \systF se distingue du lambda calcul simplement typé par l'ajout d'une opération de quantification sur les types dans les termes et les types --- cette quantification sur les types 
joue un rôle déterminant (!) dans notre traitement des déterminants.

Les types sont définis inductivement à partir de types de base: 
\begin{itemize} 
\item Types de base: 
\begin{itemize} 
\item $\ttt$ \emph{les valeurs de vérité}, $\event$ les \emph{événements}, 
\item des types constants $\eee_i$ en grand nombre correspondant aux \emph{différentes sortes d'individus}, 
\item des \emph{variables de type}, notées par des lettres grecques (issues d'un ensemble dénombrable $P$)
\end{itemize}
\item Lorsque $T$ est un type et  $\alpha$ une variable de type, qui peut ou non  apparaître 
dans $T$, $\Pi \alpha.\ T$ est un type (dit polymorphe).  
\item Lorsque $T_1$ et $T_2$ sont des types, $T_1\fl T_2$ est aussi un type. 
\end{itemize}

Pour définir les termes, on se donne une infinité dénombrable de variables de chaque type, ainsi que, pour chaque type, des constantes en nombre fini (possiblement aucune): 
\begin{itemize} 
\item Une variable de type  $T$ \cad $x:T$ (ce qu'on écrit aussi $x^T$) est un  \emph{terme} de type $T$.
\item Une constante de type  $T$ \cad $c:T$ (ce qu'on écrit aussi $c^T$) est un  \emph{terme} de type $T$.
\item $(f\ \tau)$ est un terme de type $U$ quant $\tau$ est de type $T$ et   $f$ de type $T\fl U$. 
\item $\lambda x^T\!\!.\ \tau$ est un terme de type  $T\fl U$ si  $x$ est une variable de type $T$, et  $\tau$ un terme de type $U$.  
\item $\tau \{U\}$ est un terme de type $T[U/\alpha]$
quand $\tau:\Pi \alpha.\ T$, et $U$ est un type. 
\item $\Lambda \alpha. \tau$ est un terme de type $\Pi \alpha. T$
quand $\alpha$ est une variable de type  $\tau:T$ sans occurrence de $\alpha$ dans le type d'une variable libre. 
\end{itemize}

Lorsque les constantes sont celles d'une logique multisorte 
(connecteurs usuels $\ttt\fl\ttt\fl\ttt$, quantificateurs $\exists,\forall:(e_i\fl\ttt)\fl\ttt,\ldots$, constantes $Fido: ani,\ regarde:{ani}\fl \eee\fl \ttt$) ce système est appelé  $\ltyn$. 

Les réductions pour $\lambda$ et $\Lambda$ sont définies de manière similaire.
\begin{itemize} 
\item $(\Lambda \alpha. \tau) \{U\}$  se réduit en $\tau[U/\alpha]$ (rappelons que $\alpha$ et $U$ sont des types). 
\item $((\lambda x^U. \tau^T)^{U\fl T} u^U):T$ se réduit en $\tau[u/x]$ (réduction habituelle, $u$ est une terme de même type $U$ que la variable $x$). 
\end{itemize} 
La normalisation du système \systF\ a 
une conséquence heureuse pour notre modèle sémantique:  si les constantes (du $\lambda$-calcul) correspondent au langage $L$ multisorte d'une logique d'ordre $n$ (opérations logiques, prédicats, fonctions et constantes), tout terme normal de type $\ttt$ 
correspond à une formule de $L$. 

Donnons l'organisation générale de notre modèle de la sémantique compositionnelle:
\begin{description} 
\item[le $\lambda$-calcul du second ordre,] le système \systF\  sert à assembler les formules logiques partielles contenues dans le lexique (il remplace le $\lambda$-calcul simplement typé utilisé par Montague) 
\item[la logique d'ordre supérieur multisorte] dans laquelle s'expriment  les représentations sémantiques
(elle remplace la logique d'ordre supérieur utilisée par Montague, ainsi que ses variantes réifiées du premier ordre: les nombreuses sortes $\eee_i$ sont les types de base qui gèrent les restrictions de sélection). 
\end{description}

Afin d'illustrer l'utilité de la quantification sur les types, donnons un exemple avec une coprédication qui fait intervenir une conjonction polymorphe, donnée en (\ref{etpolymorphe}). 
Cette unique conjonction permet, chaque fois que l'on a   deux prédicats $P^{\alpha\fl \ttt}$, $Q^{\beta\fl \ttt}$ portant sur des entités de sortes respectives 
 $\alpha$ et  $\beta$, ainsi que des transformations $f^{\xi\fl\alpha}$ et $g^{\xi\fl\beta}$ du type $\xi$ dans $\alpha$
 et dans  $\beta$ de dire que les images d'un objet $x^\xi$ de type $\xi$ ont les propriétés 
 $P^{\alpha\fl \ttt}$ et  $Q^{\beta\fl \ttt}$.

\begin{figure} 
$$ 
\begin{array}{l|l|rl} 
\mbox{mot} & \mbox{$\lambda$-terme principal} & \multicolumn{1}{l}{\mbox{$\lambda$-termes optionnels}} & \mbox{rigide/flexible}\\ \hline 
\hline 
Liverpool & liverpool^T & Id_T:T\fl T &\flex \\ 
& & t_1:T\fl F &\rig \\ 
& & t_2:T\fl P &\flex \\ 
& & t_3:T\fl Pl &\flex\\ 
\hline 
vaste & vaste:Pl\fl\ttt & \\ 
\hline 
a\_vot\acute{e} & a\_vot\acute{e}:P\fl\ttt& \\
\hline 
a\_gagn\acute{e} & a\_gagn\acute{e}:F\fl\ttt&\\  
\end{array}
$$
où les types de base sont définis comme suit $T$ (ville), $Pl$ (lieu), $P$ (gens), $F$ (club). 
\caption{Un exemple de lexique}
\label{lexicon} 
\end{figure} 

\begin{exe}
\ex
\begin{xlist} 
\ex \textbf{Liverpool} est \textsc{vaste} et \textsc{a\_voté}. 
\ex \label{livsemterm} $\Land \{Pl\} \{P\} (est\_vaste)^{Pl\fl \ttt} (a\_vot\acute{e}))^{P\fl\ttt} \{T\} Liverpool^T (t_3^{T\fl Pl}) (t_2^{T\fl P})$ 
\ex \label{etpolymorphe}  $\Land =\Lambda \alpha \Lambda \beta
\lambda P^{\alpha \fl \ttt} \lambda Q^{\beta\fl \ttt} 
 \Lambda \xi \lambda x^\xi 
 \lambda f^{\xi\fl\alpha} \lambda g^{\xi\fl\beta}.\ 
(\textrm{and}^{\ttt\fl\ttt\fl\ttt} \ (P \ (f \ x)) (Q \ (g \  x))) 
$ 
\ex \label{livsemfinale} 
$(\textrm{and} \ (est\_vaste{Pl\fl\ttt} \ (t_3^{T\fl Pl} \ Liverpool^T)) (a\_vot\acute{e}^{Pl\fl\ttt} \ (t_2^{T\fl P} \  Liverpool^T)))$
\end{xlist} 
\end{exe} 

Cet exemple s'analyse au moyen des deux transformations $(t_3^{T\fl Pl})$ et $(t_2^{T\fl P})$
celle d'une ville $T$ en un lieu $Pl$ et celle d'une ville en habitants $P$.
Aucune des deux n'étant rigide, on peut les utiliser toutes les deux, et le 
$\lambda$-terme sémantique de la phrase est donné en (\ref{livsemterm}). On remarquera les spécialisations de types: $\alpha:=Pl$, 
 $\beta:=P$ et $\xi:=T$. 
Après réduction  on obtient comme espéré (\ref{livsemfinale}).  
La même situation avec $a\_vot\acute{e}$ et $a\_gagn\acute{e}$ serait impossible car la transformation d'une ville en club est \emph{rigide}, elle exclut celle de la ville en tant qu'habitants.

\section{Des termes typés pour prédicats et déterminants}

\subsection{Prédicats et types} 
\label{predicatstypes} 

Un déterminant ``classique" s'applique à un prédicat, voire à deux lorsqu'il s'agit d'un quantificateur généralisé, pour donner une proposition. Un opérateur de Hilbert se combine avec un prédicat pour donner un terme. 
La modélisation du déterminant, avec ou sans opérateurs de Hilbert,  est donc intimement liée à celle du prédicat.
Dans un système multisorte et typé comme $\ltyn$,  il nous 
 faut décider quels sont les types des prédicats: usuellement, un prédicat a pour type $\eee\fl\ttt$, mais en présence des innombrables types $\eee_i$ qui se partagent le rôle traditionnellement dévolu à $\eee$ , que faire? Faut-il autoriser des prédicats à avoir un domaine autre que $\eee$?  Un prédicat comme le nom commun \ma{chat} est il une propriété du type des \ma{animaux} s'il y en a un, ou est-il une propriété de toutes les entités, propriété  qui serait fausse en dehors des \ma{animaux}? 
 
Cette question est moins embarrassante qu'il n'y paraît car on peut passer d'un choix à un autre. 
En effet, un prédicat défini sur un type $\eee_i$ différent de $\eee$ (le type de toutes les entités),   comme $P^{\eee_i\fl\ttt}$ s'étend en un  $\overline{P}^{\eee\fl\ttt}$ sans difficulté, en disant qu'il est faux en dehors de $\alpha$. Réciproquement, un prédicat comme $chat$ défini sur un type d'entités $\eee_i$ (par exemple au type $ani$ des  \ma{animaux}) peut être restreint à 
tout sous type de $\eee_i$. Evidemment, un prédicat comme $chat$ restreint à un sous ensemble strict  de l'ensemble où il est vrai (par exemple au type $siamois$) et ensuite étendu à $\eee$ puis restreint aux animaux ($ani$) ne redonnera pas le  prédicat initial, car l'extension est définie uniformément sur tous les types et les prédicats comme étant fausse à l'extérieur du domaine considéré. 
Ainsi, lorsque  $\beta$ ne contient pas tous les $x:\alpha$ satisfaisant $P$ on a $\overline{(P^{\alpha\fl\ttt}|_{\beta})} |_\alpha \neq P$. Les comportements de la restriction et de l'extension du domaine d'un prédicat se définissent aisément dans un modèle ensembliste. 

On peut aussi se demander si un type définit un prédicat.  Si $ani$ est le  type des animaux, y a-t-il un prédicat \ma{être un animal}? Et si oui, quel est son domaine? Etant donné un type $\alpha$ il est difficile de dire quel type $\beta$ contenant $\alpha$ est un bon candidat pour le domaine du prédicat \emph{être de type $\alpha$}: aussi prendrons nous pour prédicat associé au type $\alpha$ le prédicat $\widehat \alpha$ de type $\eee\fl\ttt$

On voit que le  type des prédicats, 
est très lié aux types de base disponibles, qu'aucun chercheur du domaine ne prétend avoir définitivement  identifiés.  Voici quelques réponse possibles:  

\begin{itemize}
\item Un seul type  $\eee$ pour toutes les entités, ce qui exclut toute considération lexicale. 
\item \`A l'opposé de la solution minimale que nous venons d'évoquer,  il y a une solution maximale selon laquelle toute formule à une variable libre définit un type. Il n'est pas sûr qu'un tel système soit bien fondé puisque 
les formules sont définies au moyen des types. 
 \item 
\citet{asher-webofwords} propose d'utiliser un petit nombre de type de base qui correspondraient à des classes ontologiques simples \ma{événement, objet physique, contenu informationnel, humain,...} correspondant aux restrictions de sélection que l'on rencontre dans la langue. 
\item 
\citet{Luo2012lacl} propose d'utiliser tous les noms communs.
\item 
Nous n'avons pas d'avis tranché sur la question, 
mais nous faisons remarquer à la solution précédente, qu'il faut sans doute ajouter aux noms communs des types pour les propositions et pour les verbes d'action, puisqu'on quantifie aussi sur ce type d'objet:
\begin{exe}
\ex 
\begin{xlist}
\ex Elle voudrait qu'il croit en \textsc{tout ce qu'elle lui dit}.
\ex Il a fait \textsc{tout ce qu’il a pu} et il n’a même pas voulu être payé. 
\end{xlist} 
\end{exe} 
\end{itemize} 

\subsection{Des termes typés pour les déterminants}

Nous proposons que les déterminants indéfinis soit modélisés par une constante $\epsilon$ de type $\Pi\alpha.\ (\alpha\fl\ttt)\fl\alpha$ --- le $\epsilon$ de Hilbert adapté au cadre typé et multisorte. 
Nous voyons donc l'article indéfini comme un $\epsilon$ polymorphe, qui se spécialise au type $\{\eee_i\}$ pour s'appliquer à un prédicat $P$ de type $\eee_i\fl\ttt$: il produit  un objet du type $\eee_i$. 
Considérons l'exemple  suivant, très simple et inventé,  afin d'illustrer notre traitement des déterminants 
($ani$ désigne le types des animaux): 

\begin{exe}
\ex \begin{xlist}
\ex Un chat dort (sous ta voiture). 
\ex \label{chattermun} terme pour \ma{un}: $\epsilon:\Pi\alpha.\ ((\alpha\fl\ttt)\fl\alpha)$
\ex \label{chatsynt} syntaxe: $((un \fl chat) \leftarrow dort)$ 
\ex \label{chatsem} sémantique: $dort (un\ chat)$ 
\ex \label{chattermsem} $(\lambda x.\ dort^{ani\fl \ttt}(x)) (\epsilon^{\Pi\alpha.\ ((\alpha\fl\ttt)\fl\alpha)} chat^{ani\fl \ttt})$ 
\ex \label{chattermsemac} $(\lambda x.\ dort(x)) (\epsilon^{\Pi\alpha.\ ((\alpha\fl\ttt)\fl\alpha)} \{ani\} chat^{ani\fl \ttt})$ 
\ex \label{chattermsemred}  $dort^{ani\fl \ttt}  (\epsilon^{\Pi\alpha.\ ((\alpha\fl\ttt)\fl\alpha)} \{ani\} chat^{ani\fl \ttt}):\ttt$ 
\ex \label{chattermsempresup} $chat(\epsilon^{\Pi\alpha.\ ((\alpha\fl\ttt)\fl\alpha)} \{ani\} chat^{ani\fl \ttt}):\ttt$ 
\end{xlist} 
\end{exe}

La syntaxe fournit un arbre binaire qui indique quel constituant s'applique à l'autre (\ref{chatsynt}). 
Comme  la sémantique de \ma{chat} est un prédicat qui s'applique aux entités de type \ma{animal}, on obtient le $\lambda$-terme sémantique (\ref{chattermsem}). 
Comme le type du $\lambda$-terme  sémantique associé à  \ma{un} commence par $\Pi \alpha$ (\ref{chattermun}), la 
variable de type $\alpha$ doit  s'instancier en $ani$, pour que le terme soit bien typé. 
On le voit en (\ref{chattermsemac}) avec l'application du terme de \ma{un} au \emph{type} $ani$:  
$\epsilon^{\Pi\alpha.\ ((\alpha\fl\ttt)\fl\alpha)} \{ani\}$. 
Ce terme de type $(ani\fl \ttt)\fl ani$,  est appliqué à $chat$ de type 
$ani\fl \ttt$, donnant 
la sémantique de \ma{un chat} 
qui est de type $ani$. 
Ce  groupe nominal est le sujet du groupe verbal 
\ma{dort (sous ta voiture)}, prédicat qui s'applique à une entité de type  \ma{animal}. Le terme complet (\ref{chattermsemac}) est bien typé de type $\ttt$ 
et il se réduit 
en  (\ref{chattermsemred})
  --- ce qui, sous des conditions de non vacuité très naturelles,  peut se comprendre comme  $\exists x:ani\quad dort(x)$. 
  
C'est plutôt satisfaisant, mais rien ne dit que \ma{un chat} ait la propriété d'être un chat!  Le calcul de la sémantique de \ma{un chat} ne produit pas cela. 
Cependant nous savons  $P(\epsilon_x. P(x))\equiv \exists x\ P(x)$. 
Aussi l'énonciation de  \ma{un} chat dans le sens d'un chat particulier (et non d'un chat générique)  nous conduit-elle à ajouter la présupposition  $chat(un\ chat)$ (\cad $chat(\epsilon_x. chat(x))$) --- 
le $\lambda$-terme correspondant est donné en \ref{chattermsempresup}.  On notera que $un\ chat$ étant de type \ma{animal} le prédicat \ma{chat} peut effectivement s'y appliquer. 
Si la présupposition $F(\epsilon_x. F(x))$ est introduite a priori  sans avoir 
rencontré \ma{un F} cela revient à affirmer que la propriété $F$ est satisfaite par au moins un individu.
On peut discuter des $F$ pour lesquelles cette présupposition est fondée.  

Si \ma{chat} est un type et non une propriété, comme dans \citet{Luo2012lacl}, comment faire? 
On change le type \ma{chat}  en la propriété correspondante $\widehat{chat}$ comme expliqué au paragraphe (\ref{predicatstypes}), puis on procède comme ci-dessus. 
\footnote{Une variante: si  \ma{chat} est un type et non une propriété, on peut aussi utiliser pour \ma{un} le terme   $\epsilon':\Pi\alpha.\ \alpha$ 
(une constante de type  $\bot$ ne peut nuire à la cohérence du système).  Si on applique cette constante au type \ma{chat} obtient alors \ma{un chat} de type \ma{chat} sans ajouter de présupposition. Comme le fait pertinemment remarquer \cite{asher-webofwords} une déclaration de type $x:T$ est une forme de présupposition, car il est quasi impossible de la nier. On peut appliquer le prédicat \ma{dort} à ce chat, puisque l'inclusion $chat\subset ani$ est une  transformation lexicale.}

On peut traiter de la même manière les articles définis, comme le fait von Heusinger: seul le calcul de la référence sera différent. Tandis que l'article indéfini requiert un nouvel élément, l'article défini choisit au contraire un élément déjà présent en contexte. 
L'approche permet aussi de traiter l'interprétation des pronoms de type E de Evans. Le fait que les termes génériques soient typés n'y change rien. 
Pour interpréter les anaphores comme le \ma{il} de l'exemple (\ref{hommebis}) il suffit de recopier le terme sémantique associé à l'antécédent de \ma{il}.

La quantification universelle, peu étudiée dans un cadre typé,  est extraordinairement simple: elle correspond au terme générique $\tau_x.P(x)$ (c.f. section \ref{Hilbert}) bien plus facile à interpréter que $\epsilon_x.P(x)$.
L'élément $\tau_x.P(x)$ est celui des démonstrations mathématiques:  un objet qui, par rapport à $F$ n'a pas de propriété particulière. Ainsi, lorsque 
$\tau_x.\ P(x)$ a la propriété $P$ tous les objets l'ont. 

\section{Implémentation}

Le traitement que nous proposons des déterminants et des quantificateurs ne nécessite pas 
de modifier l'organisation de l'analyseur syntaxique et sémantique du français Grail.  
L'extension au système \systF requise par la sémantique lexicale  a déjà été testée, 
du moins sur les parties du lexique dotées d'un typage avec plusieurs sortes, afin de vérifier que $\epsilon$ s'instancie convenablement et que la réduction produit les formules quantifiées attendues. 
La grammaire a été acquise sur corpus, mais à l'heure actuelle nul ne sait comment acquérir automatiquement les lexiques sémantiques convenablement typés que nous utilisons. 
\cite{moot10grail,moot10semi}

Du point de vue syntaxique, les déterminants et quantificateurs ont ici une catégorie plus simple que d'habitude, et surtout ils n'en n'ont qu'une: $gn / n$ suffit alors qu'habituellement il en faut une par position syntaxique.  C'est à rapprocher de nos travaux sur l'interprétation sémantique de la grammaire générative \cite{ALR2010Lambek}

L'implémentation de \citet{moot10grail} utilise la 
$\lambda$-DRT plutôt que le $\lambda$-calcul  pour calculer les représentations sémantiques,
afin de de mieux suivre la structure discursive et aussi de mettre en oeuvre le lien entre opérateurs de Hilbert et liage dynamique des variables existentielles en DRT \cite{Heusinger2004}. 
Techniquement la $\lambda$-DRT change peu de choses à notre propos mais aurait nécessité beaucoup de rappels. 

\section{Conclusion} 

Ce travail pose à la fois des questions d'analyse sémantique automatique et de logique. 

La portée des quantificateurs à la Hilbert doit être discutée ainsi que le lien avec le calcul des prédicats dynamiques. Les opérateurs de Hilbert autorisent des formules sous spécifiées et incluent un liage dynamique: correspondent-ils à ceux couramment utilisé en sémantique? 

Les pluriels ont aussi un  lien avec la quantification, et nous n'en avons pas parlé, en dépit d'un premier travail de \citet{MootRetore2011coconat} dans ce même cadre.  Cette question de sémantique est assurément intéressante, elle rejoint des idées anciennes sur les pluriels avec des opérateurs qui gèrent les lectures distributives ou collectives.

Déterminer les types de base est une question importante, surtout en pratique. 
Peut-être n'y a-t-il pas de réponse en général: leur choix dépend des 
restrictions de sélection dont on souhaite rendre compte,  \cad du type d'informations attendues.
Par exemple, pour extraire les itinéraires d'un corpus de récits de voyages du XIXe,  des types d'entités spatiales et temporelles se dégagent naturellement. 
\cite{LMRS2012taln}

Notre travail pose  également des questions d'acquisition, d'une part des types de base, mais aussi des $\lambda$-termes sémantiques, non pour les déterminants qui sont connus ainsi que leur types et termes sémantiques mais pour les autres mots, noms, verbes, adjectifs.  Quels sont leurs termes sémantiques?   Avec quels types de base sont-ils écrits? L'analyseur a besoin de ces informations pour 
calculer les représentations sémantiques assez fines utilisées ici.

L'interprétation des formules avec $\epsilon$  qui ne sont pas équivalentes à des formules usuelles reste mystérieuse. \cite{epsilonIEP,espilonURL}
\`A ce jour, seule une interprétation  très complexe et possiblement erronée a été proposé par \citet{Asser1957} tandis que  \citet{Heusinger2004} a défini une interprétation assez intuitive qui dépend du contexte discursif,  mais qui perd  l'équivalence avec la quantification usuelle.  Peut-on proposer mieux? 

Tant pour la compréhension de la logique sous-jacente que pour l'organisation du modèle d'analyse sémantique automatique, nous souhaiterions mieux comprendre l'interaction entre les types utilisés pour la composition des sens  et les prédicats de la logique multi-sorte où s'exprime le sens. 


\bibliographystyle{plain}
\bibliography{bigbiblio}

\begin{thebibliography}{10}

\bibitem{ALR2010Lambek}
Maxime Amblard, Alain Lecomte, and Christian Retor{\'{e}}.
\newblock Categorial minimalist grammars: From generative grammar to logical
  form.
\newblock {\em Linguistic Analysis}, 36(1--4):273--306, 2010.

\bibitem{asher-webofwords}
Nicholas Asher.
\newblock {\em Lexical Meaning in context -- a web of words}.
\newblock Cambridge University press, 2011.

\bibitem{Asser1957}
Gunter Asser.
\newblock Theorie der logischen auswahlfunktionen.
\newblock {\em Zeitschrift f{\"u}r Mathematische Logik und Grundlagen der
  Mathematik}, 1957.

\bibitem{espilonURL}
Jeremy Avigad and Richard Zach.
\newblock The epsilon calculus.
\newblock In Edward~N. Zalta, editor, {\em The Stanford Encyclopedia of
  Philosophy}. Center for the Study of Language and Information, 2008.
\newblock \url{http://plato.stanford.edu/}.

\bibitem{BMRjolli}
{C}hristian {B}assac, {B}runo {M}ery, and {C}hristian {R}etor{\'e}.
\newblock {T}owards a {T}ype-{T}heoretical {A}ccount of {L}exical {S}emantics.
\newblock {\em {J}ournal of {L}ogic {L}anguage and {I}nformation},
  19(2):229--245, April 2010.
\newblock \url{http://hal.inria.fr/inria-00408308/}.

\bibitem{Bos2008STEP2}
Johan Bos.
\newblock Wide-coverage semantic analysis with boxer.
\newblock In Johan Bos and Rodolfo Delmonte, editors, {\em Semantics in Text
  Processing. STEP 2008 Conference Proceedings}, Research in Computational
  Semantics, pages 277--286. College Publications, 2008.

\bibitem{Corblin2004det}
Francis Corblin, Ileana Comorovski, Brenda Laca, and Claire Beyssade.
\newblock Generalized quantifiers, dynamic semantics, and french determiners.
\newblock In Francis Corblin and Henri{\"e}tte de~Swart, editors, {\em Handbook
  of French Semantic}, chapter~1, pages 3--22. {CSLI} {P}ublications, 2004.

\bibitem{EgliHeusinger1995}
Urs Egli and Klaus von Heusinger.
\newblock The epsilon operator and {E}-type pronouns.
\newblock In Urs Egli, Peter~E. Pause, Christoph Schwarze, Arnim von Stechow,
  and G{\"o}tz Wienold, editors, {\em Lexical Knowledge in the Organization of
  Language}, pages 121--141. Benjamins, 1995.

\bibitem{Evans77pronouns}
Gareth Evans.
\newblock Pronouns, quantifiers, and relative clauses (i).
\newblock {\em Canadian Journal of Philosophy}, 7(3):467--536, 1977.

\bibitem{geach1962reference}
Peter~Thomas Geach.
\newblock {\em Reference and generality: an examination of some medieval and
  modern theories}.
\newblock Contemporary philosophy. Cornell University Press, 1962.

\bibitem{HBvol2}
David Hilbert and Paul Bernays.
\newblock {\em Grundlagen der Mathematik. Bd. 2.}
\newblock Springer, 1939.
\newblock Traduction fran{\c c}aise de F. Gaillard, E. Guillaume et M.
  Guillaume, L'Harmattan, 2001.

\bibitem{LMRS2012taln}
Ana{\"\i}s Lefeuvre, Richard Moot, Christian Retor{\'e}, and No{\'e}mie-Fleur
  Sandillon-Rezer.
\newblock Traitement automatique sur corpus de r{\'e}cits de voyages
  pyr{\'e}n{\'e}ens : Une analyse syntaxique, s{\'e}mantique et temporelle.
\newblock In {\em Traitement Automatique du Langage Naturel, TALN'2012},
  volume~2, pages 43--56, 2012.

\bibitem{Luo2011lacl}
Zhaohui Luo.
\newblock Contextual analysis of word meanings in type-theoretical semantics.
\newblock In Sylvain Pogodalla and Jean-Philippe Prost, editors, {\em LACL},
  volume 6736 of {\em LNCS}, pages 159--174. Springer, 2011.

\bibitem{Luo2012lacl}
Zhaohui Luo.
\newblock Common nouns as types.
\newblock In Denis B{\'e}chet and Alexander~Ja. Dikovsky, editors, {\em LACL},
  volume 7351 of {\em Lecture Notes in Computer Science}, pages 173--185.
  Springer, 2012.

\bibitem{moot10semi}
Richard Moot.
\newblock Semi-automated extraction of a wide-coverage type-logical grammar for
  {French}.
\newblock In {\em Proceedings of Traitement Automatique des Langues Naturelles
  (TALN)}, Montreal, 2010.

\bibitem{moot10grail}
Richard Moot.
\newblock Wide-coverage {French} syntax and semantics using {Grail}.
\newblock In {\em Proceedings of Traitement Automatique des Langues Naturelles
  (TALN)}, Montreal, 2010.

\bibitem{MPR2011taln}
Richard Moot, Laurent Pr{\'e}vot, and Christian Retor{\'e}.
\newblock Un calcul de termes typ{\'e}s pour la pragmatique lexicale ---
  chemins et voyageurs fictifs dans un corpus de r{\'e}cits de voyages.
\newblock In {\em Traitement Automatique du Langage Naturel, TALN 2011}, pages
  161--166, Montpellier, France, June 2011.

\bibitem{MootRetore2011coconat}
Richard Moot and Christian Retor{\'e}.
\newblock Second order lambda calculus for meaning assembly: on the logical
  syntax of plurals.
\newblock In Reinhard Muskens, editor, {\em Coconat: Conference on Computing
  Natural Reasoning}. University of Tilburg, December 2011.
\newblock \url{http://hal.inria.fr/hal-00650644}.

\bibitem{MootRetore2012lcg}
Richard Moot and Christian Retor{\'e}.
\newblock {\em The logic of categorial grammars: a deductive account of natural
  language syntax and semantics}, volume 6850 of {\em LNCS}.
\newblock Springer, 2012.
\newblock
  \url{http://www.springer.com/computer/theoretical+computer+science/book/978-3-642-31554-1}.

\bibitem{RealCoelhoRetore2013unilog}
Livy-Maria Real-Coelho and Christian Retor{\'e}.
\newblock A generative montagovian lexicon for polysemous deverbal nouns.
\newblock In {\em 4th World Congress and School on Universal Logic -- Workshop
  on Logic and linguistics.}, Rio de Janeiro, April 2013.

\bibitem{Retore2012rlv}
Christian Retor{\'e}.
\newblock Variable types for meaning assembly: a logical syntax for generic
  noun phrases introduced by "most".
\newblock {\em Recherches Linguistiques de Vincennes}, 41:83--102, 2012.
\newblock \url{http://hal.archives-ouvertes.fr/hal-00677312}.

\bibitem{Russell1905}
Bertrand Russell.
\newblock On denoting.
\newblock {\em Mind}, 56(14):479--493, 1905.

\bibitem{epsilonIEP}
Barry~Hartley Slater.
\newblock Epsilon calculi.
\newblock {\em The Internet Encyclopedia of Philosophy}, 2005.
\newblock \url{http://www.iep.utm.edu}.

\bibitem{Steedman2012scope}
Mark Steedman.
\newblock {\em Taking Scope: The Natural Semantics of Quantifiers}.
\newblock MIT Press, 2012.

\bibitem{Heusinger1997}
Klaus von Heusinger.
\newblock Definite descriptions and choice functions.
\newblock In S.~Akama, editor, {\em Logic, Language and Computation}, pages
  61--91. Kluwer, 1997.

\bibitem{Heusinger2004}
Klaus von Heusinger.
\newblock Choice functions and the anaphoric semantics of definite nps.
\newblock {\em Research on Language and Computation}, 2:309--329, 2004.

\end{thebibliography}

\end{document}